\documentclass{article}

\usepackage{algorithm}
\usepackage{algorithmic}
\usepackage{geometry}
\usepackage{subfig}
\usepackage{amsmath}    % need for subequations
\usepackage{graphics}
\usepackage{graphicx}   % need for figures
\usepackage{verbatim}   % useful for pLrogram listings
\usepackage{color}      % use if color is used in text
\usepackage{graphics}
\usepackage{pdflscape}
\usepackage{setspace}
\usepackage{epsfig}
\usepackage[parfill]{parskip}    % Activate to begin paragraphs with an empty line
\usepackage{url}
\usepackage[applemac]{inputenc}
\usepackage{nomencl}
\usepackage{amsfonts}
\usepackage{amsmath}
\usepackage{subfig}
\usepackage{multirow}
\usepackage{array}
\usepackage{colortbl}
\usepackage{hyperref}

\newcommand{\ket}[1]{|#1\rangle}

\let\oldsqrt\sqrt
\def\sqrt{\mathpalette\DHLhksqrt}
\def\DHLhksqrt#1#2{%
\setbox0=\hbox{$#1\oldsqrt{#2\,}$}\dimen0=\ht0
\advance\dimen0-0.2\ht0
\setbox2=\hbox{\vrule height\ht0 depth -\dimen0}%
{\box0\lower0.4pt\box2}}

\title{A Quantum Production Model}

\author{
    Luís Tarrataca and Andreas Wichert\\
    Department of Informatics\\
    INESC-ID  / IST - Technical University of Lisboa\\
    Portugal\\
    \small{ \bf  \{luis.tarrataca,andreas.wichert\}@ist.utl.pt} 
}

\date{}

\begin{document}

\floatname{algorithm}{Procedure}

\newcounter{DefinitionCounter}
\setcounter{DefinitionCounter}{1}

\newcounter{ContributionCounter}
\setcounter{ContributionCounter}{1}

\maketitle

\begin{abstract}

The production system is a theoretical model of computation relevant to the artificial intelligence field allowing for problem solving procedures such as hierarchical tree search. In this work we explore some of the connections between artificial intelligence and quantum computation by presenting a model for a quantum production system. Our approach focuses on initially developing a model for a reversible production system which is a simple mapping of Bennett's reversible Turing machine. We then expand on this result in order to accommodate for the requirements of quantum computation. We present the details of how our proposition can be used alongside Grover's algorithm in order to yield a speedup comparatively to its classical counterpart. We discuss the requirements associated with such a speedup and how it compares against a similar quantum hierarchical search approach.

%\keywords{quantum computation, artificial intelligence, tree search}
%\subclass{68Q05 \and 68Q12 \and 81P40}

\end{abstract}

\section{Introduction \label{sec:introduction}}

The artificial intelligence community has since its inception focused on developing algorithmic procedures capable of modeling problem solving behaviour. Typically, this process requires the ability to translate into abstract terms environmental concepts and the set of appropriate actions that act upon them. This type of knowledge enables problem-solving agents to consider the environment and the sequence of actions allowing for a given goal state to be reached. This process is also commonly referred to as reasoning \cite{luger1993}. The production system is a formalism for describing the theory of computation. The initial set of ideas for the production system is due to the influential work of Emil Post \cite{post1943}. Production system theory describes how to form a sequence of actions leading to a desired state. Production system theory also presents a computational theory of how humans solve problems \cite{anderson1995}.   Some of the best known examples of human cognition-based production systems include the General Problem Solver \cite{newell1959} \cite{newell1963} \cite{ernst1969} \cite{newell1972}, ACT \cite{anderson1983} and SOAR \cite{laird1986} \cite{laird1987}. Recently, applications of quantum computation in artificial intelligence were examined in \cite{ying2010}.

\subsection{Production systems \label{sec:productionSystems}}

\begin{figure}[ht]
\centering
\includegraphics[width=6.0cm]{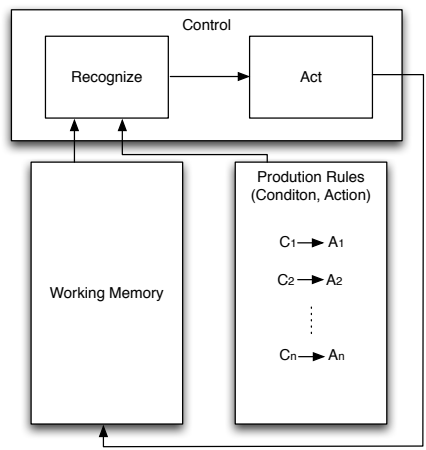}
\caption{General architecture for a production system (adapted from \cite{luger1993}). \label{fig:generalArchitectureForProductionSystems}}
\end{figure}

A production system is composed of condition-action pairs, i.e. if-then rules, which are also called productions. A computation is performed with the aid of productions through the transformation of an initial state into a desired state. The state description at any given time is also referred to as working memory. A rule is applied when the conditional part is recognized to be part of a given state. The action describes the respective problem-solving behaviour. Applying an action results in the state of the problem instance changing accordingly. On each cycle of operation, productions are matched against the working memory of facts. At any given point, more than one production might be deemed to be applicable. This subset of productions represents the conflict set. A conflict resolution strategy is then employed to this subset in order to determine an appropriate production. Finally, the action of the selected rule is carried out, changing the state of the problem instance. The operational cycle is brought to a close when a goal state is reached or when no more rules can be triggered. This general architecture is illustrated in Figure \ref{fig:generalArchitectureForProductionSystems}.

\subsection{On the power of production systems \label{sec:onThePowerOfProductionSystems}}

The first half of the twentieth century saw the beginning of the first efforts to describe intelligence in computational terms. Not surprisingly, some of the first attempts focused on developing abstract models of computation and understanding their computational limits. Some of the best known models include the Universal Turing Machine \cite{turing1936} \cite{turing1950}, Post's production system \cite{post1943}, followed closely by the thematically related Markov algorithms \cite{markov1954} and finally Church's lambda-calculus \cite{church1941}. These computational formalisms were later shown to be equivalent in power \cite{abramsky1999} \cite{martin2000} \cite{martin2001}. This power equivalency translated into an ability by all models to compute the same set of functions. Notice that this is equivalent to stating that production systems are comparable in power to a Turing machine. 

\subsection{Objectives and Problems \label{sec:objectivesAndProblems}}

In this work we propose an alternative model of quantum computation based on production system theory with a clear emphasis on problem-solving behaviour. Traditional approaches such as the quantum Turing machine are complex mechanisms oriented towards general purpose computation. A quantum production system model would be more suited to typical artificial intelligence tasks such as reasoning, inference and hierarchical search. From the onset it is possible to immediately pose some questions, namely: How should such a quantum production system model be developed? What are the requirements of quantum computation and its respective impact on the aforementioned model? How to develop the associated unitary operator? Additionally, what are the performance gains from employing quantum mechanics and how does such a proposition compare against similar strategies? Finally, are there any requirements that should be observed for those improvements? By employing such an approach we are able to (1) provide a detailed explanation of how to develop a quantum production system model; (2) assess the main differences between our proposition and its classical analog; and (3) provide an insight into better describing the power of quantum computation. However, it is not our intention to present an exact characterization of quantum computational models. Answering this question would have far-reaching consequences on complexity theory which are beyond the scope of this work.

The following sections are organized as follows: Section \ref{sec:formalDefinitions} presents the formal definitions for our proposition of a quantum production system; Section \ref{sec:classicalVsQuantumComparison} presents an assessment comparing the performance of classical production systems against our quantum proposition. We present the concluding remarks of our work in Section \ref{sec:conclusions}.

\section{Formal Definitions \label{sec:formalDefinitions}}

In this section we present a modular approach of our quantum production system proposition. Accordingly, we choose to start by introducing the set of definitions incorporating traditional production system behaviour in Section {\ref{sec:classicalProductionSystem}}. We then build on these notions to discuss reversibility requirements associated with quantum computation in Section \ref{sec:reversibleRequirementsOfAProductionSystem}. These concepts are then employed to enumerate the characteristics of a probabilistic production system in Section \ref{sec:probabilisticProductionSystem}. The probabilistic model will serve as a basis for our quantum production system which will extend those concepts in Section \ref{sec:quantumProductionSystem}.

\subsection{Classical Production System \label{sec:classicalProductionSystem}}

Any approach to a general quantum production system model needs to incorporate powerful computational abstractions, which are not bounded by input length, in a similar manner to the classical Turing machine \cite{turing1936} and its quantum counterparts. Accordingly, we choose to present the following definitions through set theory. As previously discussed each production system $S$ consists of a set of production rules $R$ and a control system $C$ alongside a working memory $W$. The following definitions embody the production system behaviour discussed in Section \ref{sec:productionSystems}.

\begin{description}

	\item[Definition \theDefinitionCounter \stepcounter{DefinitionCounter}:] Let $\Gamma$ be a finite nonempty set whose elements are referred to as symbols. Additionally, let $\Gamma^{*}$ be the set of finite strings over $\Gamma$.

	\item[Definition \theDefinitionCounter \stepcounter{DefinitionCounter}:] The working memory $W$ is capable of holding a string belonging to $\Gamma^{*}$. The working memory is initialized with a given string, who is also commonly referred to as the initial state $\gamma_{i}$.
	
	\item[Definition \theDefinitionCounter \stepcounter{DefinitionCounter}:] The set of production rules $R$ has the form presented in Expression \ref{ch:finalConsiderations:eq:productionRule}.
	
	\begin{equation}
	\{(precondition, action) | precondition, action \in \Gamma \}
	\label{ch:finalConsiderations:eq:productionRule}
	\end{equation}
	
	Each rules precondition is matched against the contents of the working memory. If the precondition is met then the action part of the rule can be applied, changing the contents of the working memory. 
	
	\item[Definition \theDefinitionCounter \stepcounter{DefinitionCounter}:] The formal definition of a production system $S$ is a tuple $(\Gamma, S_{i}, S_{g}, R, C )$ where $\Gamma, R$ are finite nonempty sets and $S_{i},S_{g} \subset \Gamma^{*}$ are, respectively, the set of initial and goal states. The control function $C$ satisfies Expression \ref{eq:productionSystemTuple}.
	
	\begin{equation}
	C : \Gamma \rightarrow R \times \Gamma \times \{h,c\}
	\label{eq:productionSystemTuple}
	\end{equation}
	
	The control system $C$ chooses which of the rules to apply and terminates the computation when a goal configuration, $\gamma_{g}$, of the memory is reached. If $C( \gamma) = (r, \gamma^{\prime}, \{h,c\})$ the interpretation is that, if the working memory contains symbol $\gamma$ then it is substituted by the action $\gamma^{\prime}$ of rule $r$ and the computation either continues, $c$, or halts, $h$. Traditionally, the computation halts when a goal state $\gamma_{g} \in S_{g}$ is achieved through a production, and continues otherwise.
	 	
\end{description}

\subsection{Reversible Requirements \label{sec:reversibleRequirementsOfAProductionSystem}}

In quantum computation, discrete state evolution of a closed system is achieved through mathematical maps known as unitary operators \cite{nielsen2000}. These maps correspond to injective and surjective functions, \text{i.e.} bijections. Bijections guarantee that every element of the codomain is mapped by exactly one element of the domain \cite{bourbaki2004}. From a computational perspective, the bijection requirement can be obtained by employing reversible computation. Classical computation is an irreversible process since at its core the use of many-to-one binary gates makes it impossible to ensure a one-to-one and onto mapping. A computation is said to be reversible if given the outputs we can uniquely recover the inputs \cite{toffoli1980a} \cite{toffoli1980b}. Irreversible computational processes can be made reversible by (1) substituting irreversible logic elements by the adequate reversible equivalents; or (2) by accounting for the information that is traditionally lost. 

The emphasis in production system theory consists in determining what state is obtained after applying a production. We employ forward chaining when moving from the conditions to the actions, \textit{i.e.} an action is applied when all the associated conditions are met. Conversely, there may be a need for determining  which state preceded the current state, \textit{i.e.} a sort of backtrace mechanism from a given state up until another state. This mechanism allowing one to reverse the actions applied and thus obtaining the associated conditions is also commonly referred to as backward chaining. Although this behaviour seems fairly simple and intuitive it is possible to immediately pose an elaborate question regarding the system's nature, namely what are the requirements associated with a reversible production system?

It is possible to adapt Bennett's original set of definitions \cite{bennett1973}  in order to describe the behaviour of a production system by a finite set of transition formulas also referred to as quadruples, in an allusion to the form of Expression \ref{eq:productionSystemTuple}. Each quadruple maps the present state of the working memory to its successor. By introducing the tuple terminology it becomes simpler to present the following set of definitions:

\begin{description}

	\item[Definition \theDefinitionCounter \stepcounter{DefinitionCounter}:]  A production system can be perceived as being deterministic if and only if its quadruples have non-overlapping domains. 
	
	\item[Definition \theDefinitionCounter \stepcounter{DefinitionCounter}:] A production system is said to be reversible if and only if its quadruples have non-overlapping ranges.
	
	\item[Definition \theDefinitionCounter \stepcounter{DefinitionCounter}:] A reversible and deterministic production system can be defined as a set of quadruples no two of which overlap either in domain or range.
	
\end{description}

These definitions contrast with Bennett's more elaborated model where information regarding the internal states of the control unit before and after the transition, alongside tape movement with the associated reading and writing information is maintained. In order to fully understand the exact impact of such requirements lets proceed by considering a production system responsible for sorting strings composed of letters  \textit{a}, \textit{b}, \textit{c}, \textit{d}, and \textit{e} based on \cite{winston1992}. The set of production rules is presented in Table \ref{table:stringSortingProductionSystemProductionSet}. Whenever a substring of the original string matches a rule's condition the production is applicable. Applying a specific rule consists in replacing the original substring, \textit{i.e.} precondition, by the action string. The  sequence of rules that is applied when the working memory is initialized in state ``edcba'' is illustrated in Table \ref{table:stringSortingProductionSystemExample}, with the computation proceeding until the string is fully sorted. 

Bennett \cite{bennett1973} points to the fact that any irreversible computation can be made reversible by saving all the information that is typically erased. However, this reversible history needs to be saved into a resource. Reusing this resource would require the information to be erased or thrown away, merely postponing the problem. The solution relies on performing a computation, saving the intermediate information that is typically lost, and then using this information to backtrack to the original input. Since both forward and backward stages are done in a reversible manner, the overall process always preserves the original information. 

\begin{table}
\begin{center}
\begin{tabular}{| l | l | l | r c l | } \hline
Rule	& Precondition	& Action	& \multicolumn{3}{|c|}{Symbolic} \\ \hline \hline
R1 	& ba 		& ab 	& ba &$\rightarrow$& ab\\ \hline
R2	& ca			& ac 		& ca &$\rightarrow$& ac \\ \hline
R3	& da			& ad 	& da &$\rightarrow$& ad \\ \hline
R4	& ea			& ae 	& ea &$\rightarrow$& ae \\ \hline
R5	& cb			& bc 		& cb &$\rightarrow$& bc \\ \hline
R6	& db			& bd 	& db &$\rightarrow$& bd \\ \hline
R7	& eb			& be 	& eb &$\rightarrow$& be \\ \hline
R8	& dc			& cd 		& dc &$\rightarrow$& cd \\ \hline
R9 	& ec			& ce 		& ec &$\rightarrow$& ce \\ \hline
R10	& ed			& de 	& ed &$\rightarrow$& de \\ \hline
\end{tabular}
\end{center}
  \caption{Rule set for sorting a string composed of letters \textit{a}, \textit{b}, \textit{c}, \textit{d}, and \textit{e} (adapted from \cite{luger1993}).\label{table:stringSortingProductionSystemProductionSet}}
\end{table}

\begin{table}
\begin{center}
\begin{tabular}{|l|l|l|l|l|} \hline
Iteration Number	& Working Memory	& Conflict Set			& Rule Fired 	& Continue?\\ \hline \hline
0				& edcba			& $\{R1, R5, R8, R10\}$	& R1 		& continue \\ \hline
1				& edcab			& $\{R2, R8, R10\}$		& R2 		& continue \\ \hline
2				& edacb			& $\{R5, R3, R10\}$		& R3 		& continue  \\ \hline
3				& eadcb			& $\{R5, R8, R4\}$		& R4 		& continue  \\ \hline
4				& aedcb			& $\{R5, R8, R10\}$		& R5 		& continue  \\ \hline
5				& aedbc			& $\{R6, R10\}$		& R6 		& continue  \\ \hline
6 				& aebdc			& $\{R8, R7\}$			& R7 		& continue  \\ \hline
7				& abedc			& $\{R8, R10\}$		& R8 		& continue  \\ \hline
8				& abecd			& $\{R9\}$				& R9 		& continue  \\ \hline
9				& abced			& $\{R10\}$			& R10 		& continue  \\ \hline
10				& abcde			& $\emptyset$			& $\emptyset$ 	& halt  \\ \hline
\end{tabular}
\end{center}
  \caption{An example of the sequence of rules applied for sorting a string composed of letters \textit{a}, \textit{b}, \textit{c}, \textit{d}, and \textit{e}. \label{table:stringSortingProductionSystemExample}}
\end{table}

However, before undoing the computation, care has to be taken in order to ensure that the output is preserved. This requires copying the output to an output register, an operation which has to be performed reversibly. Once the output copy has been completed it is possible to proceed with the backward stage, \textit{i.e} reverse the consequences of each quadruple application. Eventually, the computation terminates, the production system returns to its original state and the result of the procedure is stored in the output medium. In Bennett's original work the reversible Turing machine is composed of three tapes, namely \cite{bennett1973}: 

\begin{itemize}

	\item working tape - where the program's input is initially stored and computation is performed in order to obtain an output which is later reversed to the original input;
	
	\item history tape - where the information that is traditionally thrown away is kept, once the program's output has been copied the history information is used in order to revert the working tape to its original state;
	
	\item output tape - where the program's output is stored.

\end{itemize}

By observing Table \ref{table:stringSortingProductionSystemExample} it is possible to see that in order to ensure that the original input is obtained, the sequence of rules leading from an initial state $\gamma_{i}$ to a goal state $\gamma_{g}$ needs to be accounted for. This sequence of rules can be used in order to ``undo'' each action. In doing so it is possible to obtain each precondition that led to a particular action being applied, up until an initial state $\gamma_{i} \in S_{i}$.  Notice that the quadruples presented in Expression \ref{eq:productionSystemTuple} effectively convey information about which production is applied when going from a certain condition to the appropriate action. Additionally, in production system theory there exists a strong emphasis on the sequence of rules leading up to a target state. This situation contrasts with the traditional interest of merely knowing the final state of the working memory. If we allow ourselves to change Bennett's original definitions of the reversible Turing machine then it becomes possible to obtain a mapping for a reversible production system. This process can be performed by requiring that 

\begin{enumerate}

	\item applying a production results in its addition to the history tape, instead of a new control-unit state. Since the quadruple and production rules are equivalent concepts we are basically storing the same transitional information employed by Bennett's model;
	
	\item once the computation halts it is necessary to copy the contents of the history tape to the output tape, this contrasts with the original copying of the working tape. In order to do so the history tape's head needs to be place at the tape's beginning. Afterwards, the copy process from the history tape to the output tape can proceed.
	
	\item upon the copying mechanism's conclusion, the output tape's head needs to be placed at the beginning. This process can be performed by shifting left the output tape until a blank symbol is found. 
	
\end{enumerate}

Table \ref{table:stringSortingReversibleProductionSystem} illustrates this set of ideas for a reversible production simple based on the string sorting production system presented earlier (Table \ref{table:stringSortingProductionSystemProductionSet} and Table \ref{table:stringSortingProductionSystemExample}). As it is possible to verify the computation proceeds normally for iteration 0 through 10,  also known as the forward computation stage. The only alteration to Bennett's model consists in adding the productions fired to the history tape. Once this stage has concluded the history tape's head needs to be properly placed at the beginning. This step is carried out in iteration 11. In this case we opted to represent the position of a tape's head by an underbar. The system then proceeds in iteration 12 by copying the contents of the history tape onto the output tape. Additionally, the output tape's head is placed at the beginning in iteration 13. The last stage of the computation consists in undoing each one of the applied productions, as illustrated from iteration 14 to 24. For this stage we opted to represent the inverse of a rule $R$ mapping a precondition $A$ into an action $B$, \textit{i.e.} $R: A \rightarrow B$, by $R^{-1}$ such that $R^{-1}: B \rightarrow A$. By inverting the rules applied we are for all purposes reversing the consequences of each associated quadruple.

%MAYBE I SHOULD THINK THIS THROUGH AGAIN, NAMELY THE PART WHERE I TRY TO ESTABLISH THE RELATIONSHIP BETWEEN THE TUPLES AND THE SEQUENCE OF RULES OF THE REVERSIBLE PRODUCTION SYSTEM

\begin{table}
\begin{center}
\tiny{
\begin{tabular}{|l|l|l|l|l|} \hline
Iteration			& Memory			& Rule 		& History Tape 										& Output Tape \\ \hline \hline
0				& edcba			& $R1$		&$\{\underline{\mbox{  }}\}$							& $\{\underline{\mbox{  }}\}$ \\
1				& edcab			& $R2$		& $\{\underline{R1}\}$								& $\{\underline{\mbox{  }}\}$ \\ 
2				& edacb			& $R3$		& $\{R1, \underline{R2}\}$							& $\{\underline{\mbox{  }}\}$ \\ 
3				& eadcb			& $R4$		& $\{R1, R2, \underline{R3}\}$ 							& $\{\underline{\mbox{  }}\}$ \\ 
4				& aedcb			& $R5$		& $\{R1, R2, R3, \underline{R4}\}$ 						& $\{\underline{\mbox{  }}\}$ \\ 
5				& aedbc			& $R6$		& $\{R1, R2, R3, R4, \underline{R5} \}$ 					& $\{\underline{\mbox{  }}\}$ \\ 
6 				& aebdc			& $R7$		& $\{R1, R2, R3, R4, R5, \underline{R6} \}$ 				& $\{\underline{\mbox{  }}\}$ \\ 
7				& abedc			& $R8$		& $\{R1, R2, R3, R4, R5, R6, \underline{R7} \}$ 			& $\{\underline{\mbox{  }}\}$ \\ 
8				& abecd			& $R9$		& $\{R1, R2, R3, R4, R5, R6, R7, \underline{R8} \}$ 		& $\{\underline{\mbox{  }}\}$ \\ 
9				& abced			& $R10$		& $\{R1, R2, R3, R4, R5, R6, R7, R8, \underline{R9} \}$ 		& $\{\underline{\mbox{  }}\}$ \\ 
10				& abcde			& $\emptyset$ 	& $\{R1, R2, R3, R4, R5, R6, R7, R8, R9, \underline{R10} \}$	& $\{\underline{\mbox{  }}\}$ \\ \hline

11				& abcde			& $\emptyset$ 	& $\{\underline{R1}, R2, R3, R4, R5, R6, R7, R8, R9, R10 \}$	& $\{\underline{\mbox{  }}\}$	\\ \hline

12				& abcde			& $\emptyset$ 	& $\{R1, R2, R3, R4, R5, R6, R7, R8, R9, \underline{R10} \}$	& $\{R1, R2, R3, R4, R5, R6, R7, R8, R9, \underline{R10} \}$	\\ \hline

13				& abcde			& $\emptyset$ 	& $\{R1, R2, R3, R4, R5, R6, R7, R8, R9, \underline{R10} \}$	& $\{\underline{R1}, R2, R3, R4, R5, R6, R7, R8, R9, R10 \}$	\\ \hline

14				& abcde			& $\emptyset$ 	& $\{R1, R2, R3, R4, R5, R6, R7, R8, R9, \underline{R10} \}$	& $\{\underline{R1}, R2, R3, R4, R5, R6, R7, R8, R9, R10 \}$	\\
15				& abced			& $R10^{-1}$ 	& $\{R1, R2, R3, R4, R5, R6, R7, R8, \underline{R9} \}$		& $\{\underline{R1}, R2, R3, R4, R5, R6, R7, R8, R9, R10 \}$	\\ 
16				& abecd			& $R9^{-1}$ 	& $\{R1, R2, R3, R4, R5, R6, R7, \underline{R8} \}$			& $\{\underline{R1}, R2, R3, R4, R5, R6, R7, R8, R9, R10 \}$	\\ 
17				& abedc			& $R8^{-1}$ 	& $\{R1, R2, R3, R4, R5, R6, \underline{R7} \}$			& $\{\underline{R1}, R2, R3, R4, R5, R6, R7, R8, R9, R10 \}$	\\ 
18				& aebdc			& $R7^{-1}$ 	& $\{R1, R2, R3, R4, R5, \underline{R6} \}$				& $\{\underline{R1}, R2, R3, R4, R5, R6, R7, R8, R9, R10 \}$	\\ 
19				& aedbc			& $R6^{-1}$ 	& $\{R1, R2, R3, R4, \underline{R5}\}$					& $\{\underline{R1}, R2, R3, R4, R5, R6, R7, R8, R9, R10 \}$	\\ 
20				& aedcb			& $R5^{-1}$ 	& $\{R1, R2, R3, \underline{R4}\}$						& $\{\underline{R1}, R2, R3, R4, R5, R6, R7, R8, R9, R10 \}$	\\ 
21				& eadcb			& $R4^{-1}$ 	& $\{R1, R2, \underline{R3}\}$							& $\{\underline{R1}, R2, R3, R4, R5, R6, R7, R8, R9, R10 \}$	\\ 
22				& edacb			& $R3^{-1}$ 	& $\{R1, \underline{R2}\}$							& $\{\underline{R1}, R2, R3, R4, R5, R6, R7, R8, R9, R10 \}$	\\ 
23				& edcab			& $R2^{-1}$ 	& $\{\underline{R1}\}$								& $\{\underline{R1}, R2, R3, R4, R5, R6, R7, R8, R9, R10 \}$	\\ 
24				& edcba			& $R1^{-1}$ 	& $\{\underline{\mbox{  }}\}$							& $\{\underline{R1}, R2, R3, R4, R5, R6, R7, R8, R9, R10 \}$	\\ \hline

\end{tabular}}
\end{center}
  \caption{Operation of a reversible production system based on the example of Table \ref{table:stringSortingProductionSystemExample} and Bennett's model for a reversible Turing machine. The underbar denotes the position of the head.   \label{table:stringSortingReversibleProductionSystem}}
\end{table}

\subsection{Probabilistic Production System \label{sec:probabilisticProductionSystem}}

Consider a production system whose control strategy chooses a rule to apply from set of production rules based on a probability distribution. This behaviour can be formalized with a simple reformulation of Expression \ref{eq:productionSystemTuple} as illustrated by Expression \ref{eq:probabilisticControlSystem}, where $C( \gamma, r, \gamma^{\prime}, d)$ represents the probability of choosing rule $r$, substituting symbol $\gamma$ with $\gamma^{\prime}$ and making a decision $d$ on whether to continue or halt the computation if the memory contains $\gamma$.

\begin{equation}
	C : \Gamma \times R \times \Gamma \times \{h,c\} - [0,1]
	\label{eq:probabilisticControlSystem}
\end{equation}

Additionally, it would have to be required that $\forall \gamma \in \Gamma$ Expression \ref{ch:finalConsiderations:eq:probabilisticControlSystemNormalization} be observed

\begin{equation}
	\sum_{ \forall(r, \gamma^{\prime}, d)  \in R \times \Gamma \times \{h,c\} } C( \gamma, r, \gamma^{\prime}, d ) = 1 
	\label{ch:finalConsiderations:eq:probabilisticControlSystemNormalization}
\end{equation}

\begin{figure}[ht]
\centering
\includegraphics[width=10.0cm]{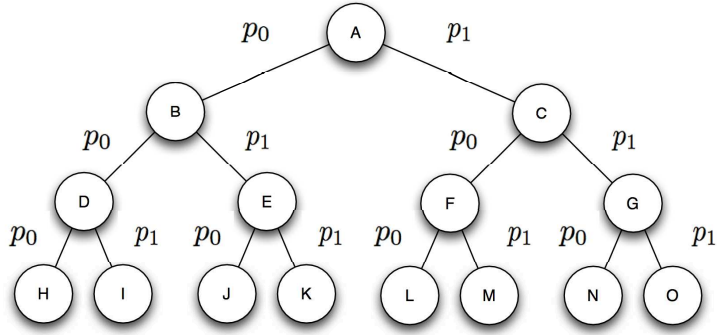}
\caption{Tree structure representing the multiple computational paths of a probabilistic production system. \label{fig:probabilisticProductionSystem}}
\end{figure}

This modification to the deterministic production system allows the control strategy to yield different states with probabilities that must sum up to 1. In such a model, a computation can be perceived has having an associated probability which is simply the multiplication of each production's probability. If the several possibilities are accounted for the overall computational process presents a tree form. Figure \ref{fig:probabilisticProductionSystem} illustrates a production system whose set of production rules is binary, \textit{i.e.} $\{p_{0}, p_{1}\}$. The root node $A$ depicts the initial state in which the working memory is initialized. Each depth layer $d$ is responsible for adding $b^{d}$ nodes to the tree, where $b$ is the branching factor induced by the production set cardinality. For this specific case $b=2$. The remaining tree nodes represent states achieved by applying the sequence of productions leading up to that specific element, \textit{e.g.} state $J$ is achieved by applying sequence $\{p_{0}, p_{1}, p_{0}\}$. 

\subsection{Quantum Production System \label{sec:quantumProductionSystem}}

A suitable model for a probabilistic production system enables a mapping between real-valued probabilities and complex-value quantum amplitudes. More specifically, the complex valued control strategy would need to behave as illustrated in Expression \ref{eq:complexValuedControlSystem} where $C( \gamma, r, \gamma^{\prime}, d)$ provides the amplitude if the working memory contains symbol $\gamma$ then rule $r$ will be chosen, substituting symbol $\gamma$ with  $\gamma^{\prime}$ and a decision $d$ made on whether to continue or halt the computation.

\begin{equation}
	C : \Gamma \times R \times \Gamma \times \{h,c\} - \mathbb{C}
	\label{eq:complexValuedControlSystem}
\end{equation}

The amplitude value provided would also have to be in accordance with Expression \ref{ch:finalConsiderations:eq:complexValuedControlSystemNormalization}, $\forall \gamma \in \Gamma$ 

\begin{equation}
	\sum_{ \forall(r, \gamma^{\prime}, d)  \in R \times \Gamma \times \{h,c\} } |C( \gamma, r, \gamma^{\prime}, d )|^{2} = 1 
	\label{ch:finalConsiderations:eq:complexValuedControlSystemNormalization}
\end{equation}

\begin{figure}[ht]%[tp]
  \centering
  	\subfloat[]	{\label{fig:irreversibleControlStrategy} \includegraphics[width=0.4\textwidth]{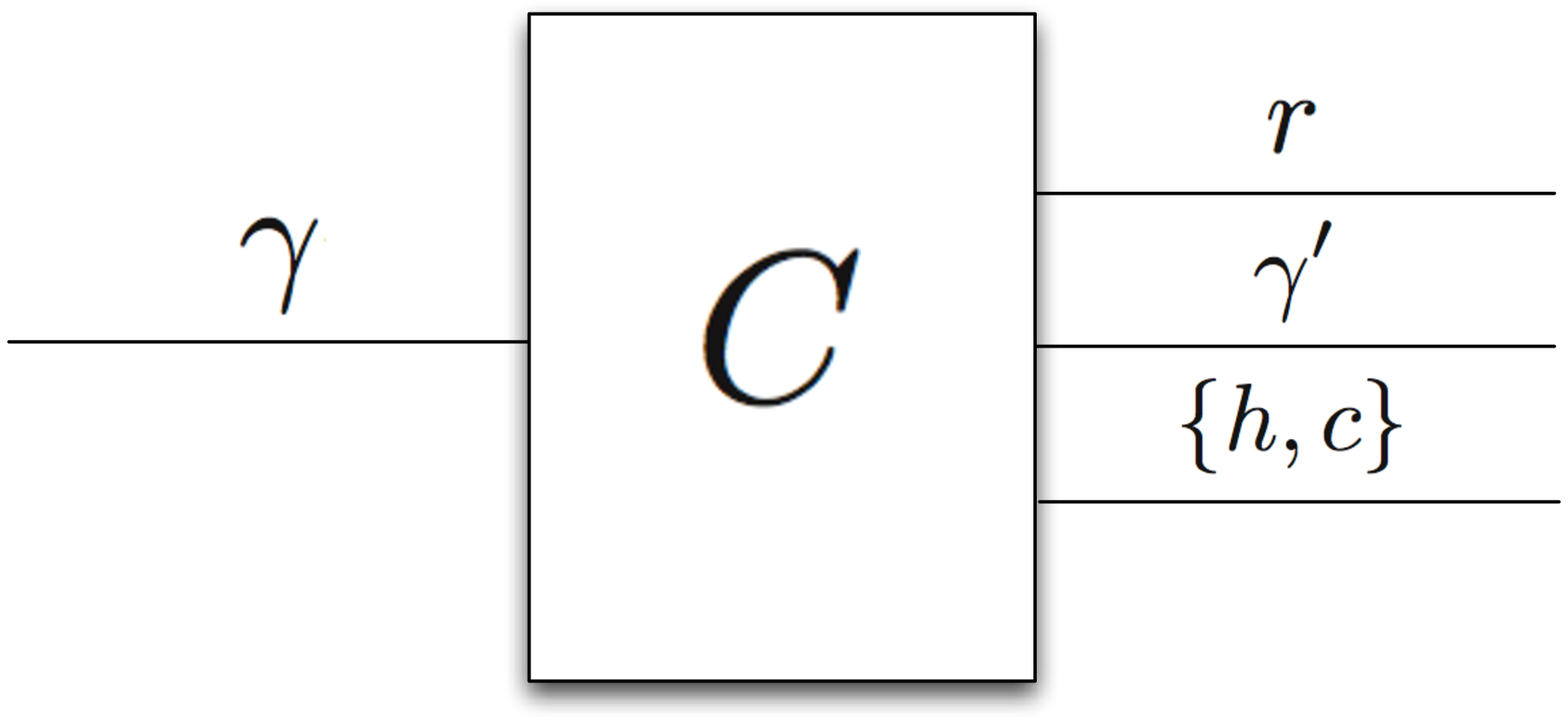}}               
 	\subfloat[]	{\label{fig:reversibleControlStrategy} \includegraphics[width=0.4\textwidth]{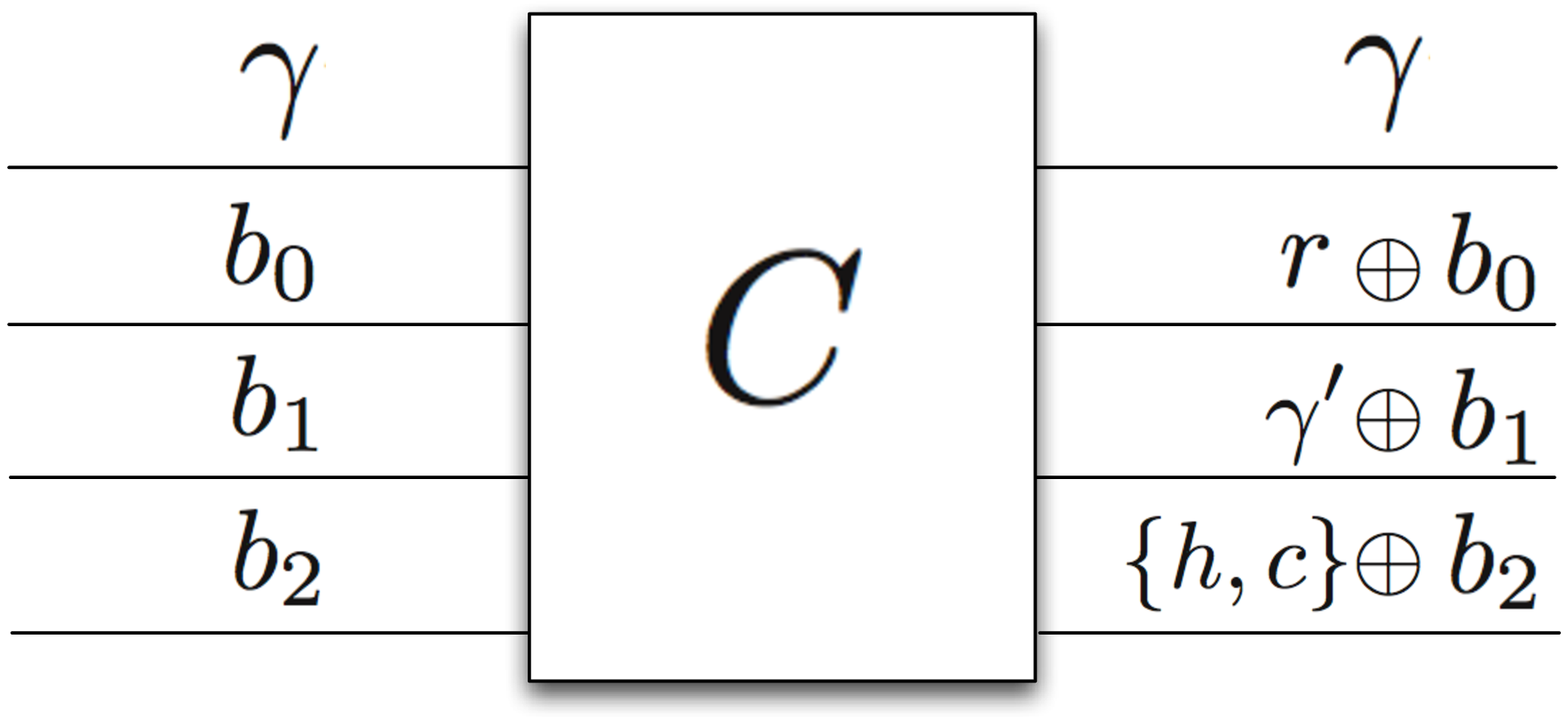}}
  \caption{ An irreversible control strategy \ref{fig:irreversibleControlStrategy} can be made reversible \ref{fig:reversibleControlStrategy} through the introduction of a number of constants and auxiliary input and output bits.}
  \label{fig:irreversibleToReversibleFunctions}
\end{figure}

Is it possible to elaborate on the exact unitary form that $C$ should take? If we were to develop a classical computational gate for calculating Expression \ref{eq:productionSystemTuple} then it would have a form as illustrated in Figure \ref{fig:irreversibleControlStrategy}. Since multiple arguments could potentially map onto the same element such a strategy would not allow for reversibility. Theoretically, any irreversible production system can be made reversible by adding some auxiliary input bits and through the addition modulo 2 operation \cite{toffoli1980b}, a process formalized in Expression \ref{eq:productionSystemReversibleCircuitBehaviour} and shown in Figure \ref{fig:reversibleControlStrategy}. Since the inputs are now part of the outputs, this mechanism allows for a bijection to be obtained.

\begin{equation}
\underbrace{(\gamma, b_{0}, b_{1}, b_{2} )}_{\text{input vector } v_{1}} \stackrel{C}{\rightarrow} \underbrace{(\gamma, r \oplus b_{0} , \gamma^{\prime} \oplus  b_{1}, \{h,c\} \oplus b_{2})}_{\text{output vector } v_{2}}
\label{eq:productionSystemReversibleCircuitBehaviour}
\end{equation}

Notice that the reversible gate can be perceived as acting upon an input vector $v_{1}$ and delivering $v_{2}$. If we adopt a linear algebra perspective alongside the Dirac notation \cite{dirac1939} \cite{dirac1981}, then such behaviour can be described as shown in Expression \ref{eq:controlStrategyUnitaryOperatorBehaviour}, where $C$ is the required unitary operator. 

\begin{equation}
C \ket{v_{1}} = \ket{v_{2}}
\label{eq:controlStrategyUnitaryOperatorBehaviour}
\end{equation}

Based on Expression \ref{eq:productionSystemReversibleCircuitBehaviour} and Expression \ref{eq:controlStrategyUnitaryOperatorBehaviour} it becomes possible to develop a unitary operator $C$. Accordingly, $C$ acts upon an input vector $v_{1}$ conveying specific information about the argument's state. From Expression \ref{eq:productionSystemReversibleCircuitBehaviour} we can verify that any input vector $\ket{v_{1}}$ should be large enough to accommodate $\gamma, b_{0}, b_{1}$ and $b_{2}$. Since $b_{0}, b_{1}$ and $b_{2}$ will be used for bitwise addition modulo 2 operations with, respectively, $r, \gamma^{\prime}$ and $\{h,c\}$, we need  to determine the appropriate dimensions for a binary encoding of these elements. Assume that:

\begin{itemize}

	\item $\alpha = \lceil \log_{2}{|\Gamma|} \rceil$, represents the number of bits required to encode the symbol set

	\item $\beta = \lceil \log_{2}{|R|} \rceil$, represents the number of bits required to encode each one of the productions;
	
	\item $\delta$ is a single bit used to encode either $h$ or $c$
	
\end{itemize}

If we employ a binary string to represent this information, then its length will be $\alpha + \beta + \delta$ bits, thus allowing for a total of $2^{\alpha + \beta + \delta}$ combinations. This information about the input's state can be conveyed in a column vector $v_{1}$ of dimension $2^{\alpha + \beta + \delta}$. The general idea being that the $m^{th}$ possible combination can be represented by placing a $1$ on the $m^{th}$ row of such a vector. These same principles are still observed by $v_{2}$. The unitary operator's responsibility relies on interpreting such information and presenting an adequate output vector $v_{2}$. The overall requirements of unitarity alongside the dimensions of input and output vectors imply that unitary operator $C$ will have dimension $2^{\alpha + \beta + \delta}\times 2^{\alpha + \beta + \delta}$.

A parallel can be established between $C$'s behaviour and the truth table concept of classical gates. Truth tables are classical mechanisms employed to describe logic gates employed in electronics. The tables list all possible combinations of the inputs alongside the respective results \cite{mano2002}. In a similar manner, we can build unitary operator $C$ by going through all possible combinations and decoding the information present in each combination. This procedure is illustrated through pseudo-code in Procedure \ref{algo:unitaryOperatorConstruction}. Lines 1-3 are employed in order to determine the required number of bits for our encoding mechanism. These values can also be used to determine the dimension $2^{\alpha + \beta + \delta} \times 2^{\alpha + \beta + \delta}$ of unitary operator $C$. This operator is initialized in line 4 as a matrix with all entries set to zero. 

The cycle \textit{for} from lines 5-15 is responsible for going through all possible combinations. Line 6 of the code obtains a string $S$ which is the binary version of decimal combination $\lambda$, represented as $\lambda_{(2)}$ to illustrate base-2 encoding. Recall from Expression \ref{fig:irreversibleToReversibleFunctions} that each input vector needs to convey information about $\gamma, b_{0}, b_{1}$ and $b_{2}$. Accordingly, for each $\lambda$ we need to parse the different elements of the string in order to determine those values. This process is illustrated through lines 7-10 which are responsible for obtaining the binary substrings. For any string $S$, $S[i,j]$ is the contiguous substring of $S$ that starts at position $i$ and ends at position $j$ of $S$ \cite{gusfield1997}. Line 11 is responsible for invoking function mapBinaryEncoding which maps substring $S_{1}$ to a symbol $\gamma \in \Gamma$. This function can be easily calculated with the help of any trivial data structure. 

Once the input symbol $\gamma$ has been determined it is possible to calculate the transition depicted in Expression \ref{eq:productionSystemTuple}. We should be careful to point out that the transition calculated in Line 12 through function $C$ should not be confused with the associated unitary operator $C$ of line 4. The next logical step consists in forming a binary string represented as $w_{(2)}$ which is simply the concatenation of elements $ S_{1}, r_{(2)} \oplus S_{2}, \gamma_{(2)}^{\prime} \oplus S_{3}$ and $d_{(2)} \oplus S_{4}$. Again, this step is done by employing the base-2 version of elements $r, \gamma^{\prime}$ and $d$. After the conclusion of line 13 we have all the information required to determine the corresponding mapping, $\lambda$ can be viewed as the decimal encoding of the input state, whilst $\omega$ can be interpreted as the new decimal state achieved. This behaviour can be adequately incorporated into the unitary operator by marking column $\lambda$ and row $\omega$ with a one, a procedure realized in line 14.

\begin{algorithm}
\caption{Pseudo code for building unitary operator $C$}
\label{algo:unitaryOperatorConstruction}

\begin{algorithmic}[1]

\STATE $\alpha = \lceil \log_{2}{|\Gamma|} \rceil$

\STATE $\beta = \lceil \log_{2}{|R|} \rceil$

\STATE $\delta = 1$

\STATE $C = \text{zeros}[2^{\alpha + \beta + \delta}, 2^{\alpha + \beta + \delta}]$

\FORALL{ integers $\lambda \in [0, 2^{\alpha + \beta + \delta}]$}

	\STATE $S$ = $\lambda_{(2)}$
	
	\STATE $S_{1}$ = $S[ 0, \alpha - 1 ]$
	
	\STATE $S_{2}$ = $S[ \alpha, \alpha + \beta - 1 ]$
	
	\STATE $S_{3}$ = $S[ \alpha + \beta, 2\alpha + \beta - 1 ]$
	
	\STATE $S_{4}$ = $S[ 2\alpha + \beta, 2\alpha + \beta + \delta - 1 ]$
	
	\STATE $\gamma$ = mapBinaryEncoding( $\Gamma$, $S_{1}$ )
	
	\STATE $C( \gamma) = (r, \gamma^{\prime}, d )$
	
	\STATE $\omega_{(2)} = S_{1}, r_{(2)} \oplus S_{2}, \gamma_{(2)}^{\prime} \oplus S_{3}, d_{(2)} \oplus S_{4}$
	
	\STATE $C_{\omega, \lambda} = 1$
	
\ENDFOR

\end{algorithmic}

\end{algorithm}

\textit{Correctness Proof:} In order to verify the correctness of Procedure \ref{algo:unitaryOperatorConstruction} we need to confirm that operator $C$ is indeed a bijective mapping. At its core a bijection performs a simple permutation of all possible input state combinations. Accordingly, for a collision to occur, \textit{i.e.} multiple arguments mapping into the same image, would require that several $\lambda$'s produced the same $\omega$. If the transition function employed in Line 12 is irreversible then it is conceivable that different $\gamma$'s may produce the same output vector $(r, \gamma^{\prime}, d )$. However, the new state $w$ besides contemplating output $(r, \gamma^{\prime}, d )$ through the addition modulo 2 elements $r_{(2)} \oplus S_{2}, \gamma_{(2)}^{\prime} \oplus S_{3}$ and $d_{(2)} \oplus S_{4}$ also takes into consideration the original input symbol $\gamma$ allowing for a differentiation of possible collision states. As a consequence, for a collision to still occur would require that function mapBinaryEncoding produced the same $\gamma$ for different binary strings. This same binary mapping behaviour can be easily avoided with proper management of an adequate data structure thus guaranteeing the correctness of such a procedure.

Notice that unitary operator $C$ is only responsible for applying a single production of the control strategy. This represents a best case scenario where a problem's solution can be found within the immediate neighbours, \textit{i.e.} those nodes that can be reached by applying a single production. However, the production system norm relies on  having to apply a sequence of rules before obtaining a solution state. Our proposition can be easily extended in order to apply multiple steps. Such an extension would require developing a logical circuit employing elementary gates $C$ alongside any necessary output redirection to the adequate inputs. Algebraically, such a procedure would require unitary operator composition acting upon the appropriate inputs, which would continue to guarantee overall reversibility. Additionally, we should emphasize that any potential unitary operator requires the ability to verify if the conditional part of a rule is met, i.e. to determine if a string contains a substring which can be achieved with simple comparison operators.

\section{Classical vs. Quantum Comparison \label{sec:classicalVsQuantumComparison}}

Deutsch described a universal model of computation capable of simulating Turing machines with inherent quantum properties such as quantum parallelism that cannot be found in their classical counterparts \cite{deutsch1985}. However, the number of computational steps required by Deutsch's model grew exponentially as a function of the simulated Turing's machine running time. Subsequently, a more efficient model for a universal quantum Turing machine was proposed in \cite{bernstein1993}. In the same work the authors questioned themselves if a quantum turing machine can provide any significant advantage over their classical equivalents. They proceeded by showing that a quantum turing machine described in \cite{deutsch1992} is capable of efficiently solving the Fourier sampling problem. However, care was also employed in order to emphasize that their result did not prove that quantum Turing machines are more powerful than probabilistic Turing machines, since the latter can sample from a distribution within $\epsilon$ total variation distance of the desired Fourier distribution \cite{bernstein1993}. Later, Shor's algorithm for fast factorization \cite{shor1994} presented further evidence on the power of quantum computation.

Naturally, the question arises: how does our quantum production system proposal fare against its classical counterpart? Namely, what do we stand to gain by applying quantum computation? And what are the requirements associated to those improvements? In order to answer these questions consider a unitary operator $C$ which is applied to an initial state $x \in S_{i}$. Additionally, assume that $C$ needs to be applied a total of $d$ times for a result to be obtained, where $d \in \mathbb{N}$ is chosen such that the computation is able to proceed until it stops. The result of applying $C$ can be represented as $g(x)$ which in production system theory can be a simple output of the productions applied. As a consequence, the quantum register employed needs to convey information about the initial state and also be large enough to accommodate for $g(x)$. We opted to represent this requirement by employing a unspecified length register $\ket{z}$. Accordingly, we can represent the initial state of the system by the left-hand side of Expression \ref{eq:quantumProductionSystem}. The right-hand side represents the result obtained after unitary evolution.

\begin{equation}
C^{d} \ket{x, z} = \ket{x, z \oplus g(x)}
\label{eq:quantumProductionSystem}
\end{equation}

\begin{figure}[ht]
\centering
\includegraphics[width=.9\textwidth]{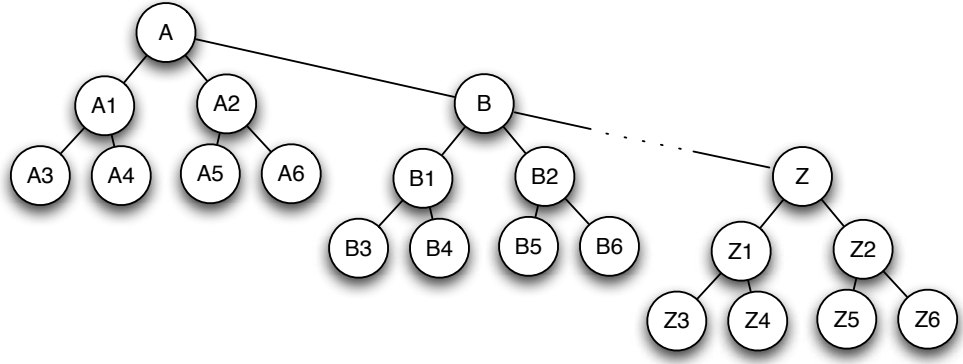}
\caption{Parallel search with $S_{i} = \{A,B, \cdots, Z\}$ and $\ket{\psi_{n}} = \frac{1}{\sqrt{|S_{i}|}} \sum_{s \in S_{i}} \ket{s}$  \label{fig:parallelTreeSearch}. The dotted lines represent the initial states belonging to superposition $\ket{\psi_{n}}$}
\end{figure}

In order to gain a quantum advantage over the classical version we need to employ the superposition principle. Accordingly, it is possible to initialize register $\ket{x}$ as a superposition, $\ket{\psi}$, of all starting states, a procedure illustrated in Expression \ref{eq:startingStateSuperposition}, where $S_{i} \subset \Gamma^{*}$ is the set of starting states. This procedure is also depicted in Figure \ref{fig:parallelTreeSearch}, where multiple binary searches are performed simultaneously, with the dotted line representing initial nodes that, for reasons of space, are not shown, but are still present in the superposition. Now consider a scenario where the production system definition only contemplates a single initial state, \textit{i.e.} $|S_{i}| = 1$. Since it is not possible to explore the high levels of parallelism provided by the superposition principle, we would therefore not have any significant advantage over the sequential procedure by applying $\ket{\psi_{n}}$. However, if the productions set cardinality is greater than one, then there exist several neighbour states which which can be employed as initial states thus circumventing the problem.

\begin{equation}
\ket{\psi} = \frac{1}{\sqrt{|S_{i}|}} \sum_{s \in S_{i}} \ket{s}
\label{eq:startingStateSuperposition}
\end{equation}

This approach differs from other strategies of hierarchical search, namely \cite{tarrataca2010} and \cite{tarrataca2011}, who, respectively, (1) evaluate a superposition of all possible tree paths up to a depth-level $d$ in order to determine if a solution is present and (2) present an hierarchical decomposition of the quantum search space through entanglement detection schemes.

The following sections are organized as follows: Section \ref{sec:theQuantumSearchAlgorithm} presents the main results on Grover's algorithm. These concepts will then be extended in Section \ref{sec:oracleExtension} in order to present a system combining our production system proposal alongside the quantum search algorithm. Finally, we will conclude in Section \ref{sec:performanceAnalysis} by discussing the performance gains achieved over the classical production system equivalent.

\subsection{The quantum search algorithm \label{sec:theQuantumSearchAlgorithm}}

Traditionally, production system theory is applied to problems devoid of an element of structure, and thus requiring the search space of all possible combinations to be exhaustively examined. The class NP consists of those problems whose possible configurations can be verified in polynomial-time. Grover's algorithm works by amplifying the amplitude of the solution states. The algorithm is able to ``mark'' a state as a solution by employing an oracle $O$ which, alongside an adequate initialization of the answer register in a superposition state, effectively flips the amplitudes of those states. This behaviour is illustrated in Expression \ref{eq:oracle}, where $\ket{x}$ and $\ket{y}$ represent, respectively, an $n$-bit query register and a single bit answer register. Function $f(x)$ simply verifies if $x$ is a solution, as formalized in Expression \ref{eq:oracleFunctionDefinition}. The quantum search algorithm \cite{grover1996} is  ideally suited for solving NP problems and allows for a quadratic speedup relatively to classical algorithms. Classical algorithms require $O(N)$ time for $N$-dimensional search spaces, whilst Grover's algorithm requires  $O(\sqrt{N})$ time, or in terms of $\ket{x}$'s dimension $O(\sqrt{2^{n}})$ time.

 \begin{equation}
	 O : \ket{ x } \ket{ y } \mapsto \ket{ x } \ket{ y \oplus f( x ) }
 \label{eq:oracle}
 \end{equation}
 
 \begin{equation}
f(x) = \left\{
		\begin{array}{ll}
		1 & \text{if $x$ is a solution}\\
		0 & \text{otherwise}
		\end{array} \right.
\label{eq:oracleFunctionDefinition}
\end{equation}

The amplification process is achieved by flipping the amplitude of the solution states and performing a inversion about the mean of the amplitudes. The overall effect of such a procedure, referred to as Grover's iterate, induces a higher probability of observing a solution when a measurement is performed over the superposition state. Grover's algorithm was experimentally demonstrated in \cite{chuang1998}. The quantum search algorithm systematically increases the probability of obtaining a solution with each iteration. Upon conclusion a measurement is performed in a quantum superposition. The superposition state represents the set of all possible results. Grover's approach sparked interest by the scientific community on whether it would be possible to devise a faster search algorithm. Subsequently, it was proved any procedure based on oracles employing total function evaluation will always require at least $\Omega(\sqrt{N})$ time \cite{bennett1997}. Grover and Radhakrishnan \cite{grover2004a} considered the speedup achievable if one was only interested in determining the first $m$ bits of a $n$ bit solution string.  In practice, their approach proceed with analysing different sections of the quantum search space. The authors prove that it is possible to obtain a speedup, however, as $m$ grows closer to $n$ the computational gains obtained disappear \cite{grover2004a}. This speedup was then improved in \cite{korepin2006} and \cite{korepin2007} and an extension to multiple solutions was presented in \cite{choi2006}.

\subsection{Oracle Extension \label{sec:oracleExtension}}

In this section we present an extension to the oracle operator employed by Grover's algorithm allowing it to be combined alongside our quantum production system proposal. As a result we need to determine what happens when two different functions $f$ and $g$ are combined into a single unitary evolution, as illustrated by Expression \ref{eq:oracleExtensionToMultipleFunctions}. In this case we opted to employ three quantum registers, namely $\ket{x}$ which is configured with the system's initial state, alongside registers $\ket{y}$ and $\ket{z}$ where, respectively, the output of functions $f(x)$ and $g(x)$ is stored. The original amplitude flipping process is a result of placing register $\ket{y}$ in the superposition state $\frac{ \ket{0} - \ket{1} }{\sqrt{2}}$. Accordingly, we need to verify if the amplitude flip still holds with the oracle formulation of Expression \ref{eq:oracleExtensionToMultipleFunctions} alongside $\ket{y}$'s superposition initialization. This behaviour is shown in Expression \ref{eq:oracleExtensionProofInitialization}. From Expression \ref{eq:oracleExtensionProofConclusion} we are able to conclude that despite the new oracle formulation the amplitude flipping continues to occur.

\begin{equation}
O \ket{x, y, z } 	= \ket{x, y \oplus f(x), z \oplus g(x)}\\
\label{eq:oracleExtensionToMultipleFunctions}
\end{equation}

\begin{align}
O \ket{x} \frac{\ket{0} - \ket{1}}{\sqrt{2}} \ket{z}&= \frac{1}{\sqrt{2}} \left( \ket{x} \ket{ f(x) } \ket{z \oplus g(x) } - O \ket{x} \ket{1 \oplus f(x) } \ket{z \oplus g(x)} \right)  \label{eq:oracleExtensionProofInitialization} \\
								       &= \begin{cases}
   									     \frac{1}{\sqrt{2}} \left( \ket{x} \ket{0} \ket{z \oplus g(x) } - \ket{x} \ket{1} \ket{z \oplus g(x)} \right) & \text{if } f(x) = 0\\
								             \frac{1}{\sqrt{2}} \left( \ket{x} \ket{1} \ket{z \oplus g(x) } - \ket{x} \ket{0} \ket{z \oplus g(x)} \right) & \text{if } f(x) = 1\\
  									     \end{cases} \nonumber \\
								       &= \begin{cases}
   									     \ket{x} \frac{\ket{0} - \ket{1}}{\sqrt{2}} \ket{z \oplus g(x) } & \text{if } f(x) = 0\\
								             \ket{x}  \frac{\ket{1} - \ket{0}}{\sqrt{2}} \ket{z \oplus g(x) } & \text{if } f(x) = 1\\
  									     \end{cases} \nonumber \\
								       &= (-1)^{f(x)} \ket{x} \frac{\ket{0} - \ket{1}}{\sqrt{2}} \ket{z \oplus g(x) } \label{eq:oracleExtensionProofConclusion}
\end{align}

\subsection{Performance Analysis \label{sec:performanceAnalysis}}

In order to proceed with our performance analysis lets consider we have a production system whose definitions are incorporated into a unitary operator $C$ combining the results of Expression \ref{eq:quantumProductionSystem} and Expression \ref{eq:oracleExtensionToMultipleFunctions}. Accordingly, $C$ will have the form presented in Expression \ref{eq:productionSystemCombiningOracleExtension}, where $\ket{x}$ is initialized with a superposition of the production system starting states. In addition we employ register $\ket{z}$ which has an unspecified length in order to accommodate for the productions applied, \textit{i.e} the output growth of function $g(x)$. By employing such a formulation for our production system $C$ we are able to employ it alongside Grover's algorithm in order to speedup the computation. In our particular case we are interested in changing $f(x)$'s definition in order to check if a goal state $s \in S_{g}$ is achieved after having applied $d$ productions. \textit{E.g.} consider that state $M$ shown in Figure \ref{fig:probabilisticProductionSystem} is a goal state, then, assuming no backtracking occurs, such state can be reached by applying productions $p_{1}, p_{0}$ and $p_{1}$. As a consequence we can express such state evolution as $C^{3} \ket{A, 0,\mathbf{\underline{0}}} = \ket{x, 1, \{p_{1},p_{0},p_{1}\}}$, where $\mathbf{\underline{0}}$ represents a vector of zeros. Function $f$ new definition is presented in Expression \ref{eq:newOracleFunctionDefinition}. The state of the system is described by a unit vector in a Hilbert space $\mathcal{H}_{2^{m}} = \mathcal{H}_{2^{n}} \otimes \mathcal{H}_{2} \otimes \mathcal{H}_{2^{p}}$.

\begin{equation}
C^{d} \ket{x, y, z} = \ket{x, y \oplus f(x), z \oplus g(x) }
\label{eq:productionSystemCombiningOracleExtension}
\end{equation}

 \begin{equation}
f(x) = \left\{
		\begin{array}{ll}
		1 & \text{if $C^{d}\ket{x} \in S_{g}$}\\
		0 & \text{otherwise}
		\end{array} \right.
\label{eq:newOracleFunctionDefinition}
\end{equation}

Grover's original speedup was dependent on superposition $\ket{\psi}$ and the associated number of possible states. More concretely, the dimension of the space spanned is dependent on the dimension of the query register $\ket{x}$ employed. However, by applying an oracle $C$ whose behaviour mimics that of Expression \ref{eq:productionSystemCombiningOracleExtension} the elements present in superposition $\ket{\psi}$ will interact with registers $\ket{y}$ and $\ket{z}$. Typically, register $\ket{y}$ is ignored when evaluating the running time, producing an overall superposition $\ket{\xi}$ which will no longer span the original $2^{n}$ possible states but $2^{n + p}$. From an algebraic perspective, the interaction process is due to the tensor product employed to describe the overall state between $\ket{x}$, $\ket{y}$ and $\ket{z}$. As a result, it is possible to pose the following question: what can be said about the growth of $\ket{z}$ and its respective impact on overall system performance?

Assume that a solution state can always be found after $d$ computational steps, either by indeed finding a goal state or by applying an heuristic function to determine an appropriate state selection. Classically, a sequential procedure would require $C = |S_{i}|\times d$ iterations, one for each initial state in need of processing. Is it possible to do any better with our proposition? Answering this question requires determining appropriate boundary conditions on the exact dimensions of $\ket{z}$ for which it is still possible to obtain a speedup over classical procedures. By employing Grover's algorithm we know that the search procedure will span the dimension of $\ket{\xi}$ which varies between $[2^{n}, 2^{n + p}]$. Accordingly, in the very unlikely best case scenario, we will be able to search all elements in $O(\sqrt{|S_{i}|})$ time. With each Grover iterate we need to apply oracle $C$ a total of $d$ times, which implies an overall number of invocations equal to $Q = \sqrt{|S_{i}|} \times d$. Therefore, a comparison is required between the classical and quantum number of iterations, respectively, $C$ and $Q$, as illustrated in Expression \ref{eq:classicalVsQuantumNumberOfIterationsV1}. The ratio presented in Expression \ref{eq:classicalVsQuantumNumberOfIterationsV1} allows us to conclude that $C$ and $Q$ differ by a factor of $\sqrt{|S_{i}|}$, effectively favoring the quantum proposal. 

\begin{equation}
\frac{C}{Q} = \frac{|S_{i}|d}{\sqrt{|S_{i}|}d} = \sqrt{|S_{i}|}
\label{eq:classicalVsQuantumNumberOfIterationsV1}
\end{equation}

 However, such a ratio does not take into account the dimension of register $\ket{z}$. Therefore, we need to determine what happens when $\ket{z}$ grows and how it affects overall performance.  Let $m$ denote the number of bits employed by registers $\ket{x}$ and $\ket{z}$, then the number of quantum iterations will be $Q = \sqrt{2^{m}} \times k$. Accordingly, Expression \ref{eq:classicalVsQuantumNumberOfIterationsV1} can be restated in terms of $m$, as depicted in Expression \ref{eq:classicalVsQuantumNumberOfIterationsV2} which effectively conveys the notion that each additional bit added to $\ket{z}$ impacts the $\frac{C}{Q}$ ratio negatively by a factor of $\frac{1}{\sqrt{2}}$. If register $\ket{z}$ is composed by $p$ bits this means that the overall decrease in performance will be $\frac{p}{\sqrt{2}} \frac{|S_{i}|}{\sqrt{2^{n}}}$, where $n$ is the number of bits required to encode the set of initial states. This result can be restated as $\frac{p}{\sqrt{2}} \sqrt{|S_{i}|}$ if we consider Grover's speedup in light of the dimension of $S_{i}$. 
 
\begin{equation}
\frac{C}{Q} = \frac{|S_{i}|}{\sqrt{2^{m}}}
\label{eq:classicalVsQuantumNumberOfIterationsV2}
\end{equation}

Additionally, we are also interested in determining when is the number of quantum iterations $Q$ smaller than the number of classical iterations $C$, as shown in Expression \ref{eq:classicalVsQuantumNumberOfIterationsV3}. 

\begin{alignat}{4}
				  & Q 		  & \quad < \quad & C \label{eq:classicalVsQuantumNumberOfIterationsV3} \\
\Leftrightarrow \quad & \sqrt{2^{m}}k & \quad < \quad & |S_{i}|k \nonumber \\
\Leftrightarrow \quad & 2^{m} 		  & \quad < \quad & |S_{i}|^{2} \nonumber \\
\Leftrightarrow \quad & m			  & \quad < \quad & \log_{2}{|S_{i}|^{2}} \label{eq:classicalVsQuantumNumberOfIterationsV4}
\end{alignat}

Expression \ref{eq:classicalVsQuantumNumberOfIterationsV4} needs to be further refined since we are trying to determine $m \in \mathbb{N}$ but the right-hand side may produce a value belonging to $\mathbb{R}$. This output is a consequence of having to deal with initial state sets $S_{i}$ whose cardinality is not a power of $2$.  Notice that the measurement of performance we have chosen, respectively, the ratio $C/Q$ will eventually be $1$ when $m = \log_{2}{|S_{i}|^{2}}$. Accordingly, if a larger number of bits is employed it effectively yields $C/Q < 1$ which will no longer translate into a speedup by the quantum version. That being the case, we choose to restrict our model to those cases where $m < \lfloor \log_{2}{|S_{i}|^{2}} \rfloor$. Furthermore, $m$ should also be large enough to contain the set of possible binary encodings of $S_{i}$, \textit{i.e.} $m \geq \lceil \log_{2}{|S_{i}|} \rceil$. The general boundary conditions are presented in Expression \ref{eq:boundaryConditionsForM}. 

\begin{equation}
\lceil \log_{2}{|S_{i}|} \rceil \leq m \leq  \lfloor \log_{2}{|S_{i}|^{2}}\rfloor
\label{eq:boundaryConditionsForM}
\end{equation}

\begin{figure}[ht]
\centering
\includegraphics[width=0.8\textwidth]{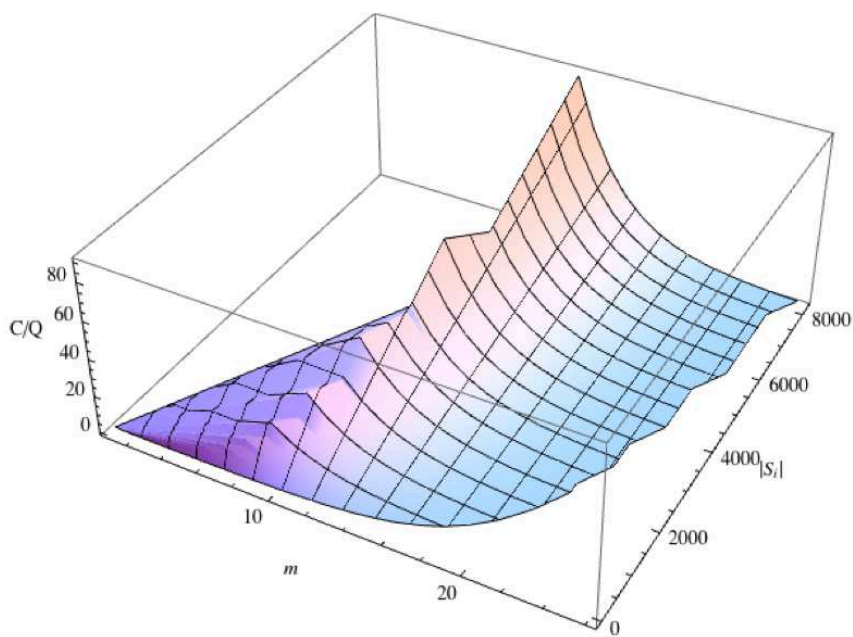}
\caption{The performance measurement ratio $C/Q$ for $|S_{i}| \in [1, 2^{13}]$ illustrating the logarithmic growth $\lceil \log_{2}{|S_{i}|} \rceil \leq m \leq  \lfloor \log_{2}{|S_{i}|^{2}}\rfloor$ alongside the $\frac{p}{\sqrt{2}} \sqrt{|S_{i}|}$ associated decrease in performance.  \label{fig: speedupVsNumberOfBitsVsNumberOfInitialStatesV1}}
\end{figure}

Figure \ref{fig: speedupVsNumberOfBitsVsNumberOfInitialStatesV1} illustrates the three-dimensional plot of Expression \ref{eq:classicalVsQuantumNumberOfIterationsV2} as a function of a number of initial nodes in the range $[1, 2^{13}]$ alongside the required boundary conditions described by Expression \ref{eq:boundaryConditionsForM}. The plot presents the characteristic ladder effect associated with employing logarithmic functions in conjunction with functions that map real domains to the integer set. As a consequence, a plateau is reached for some combinations where a number of different cardinality $S_{i}$ sets can be mapped by the same number of bits, thus presenting the same $C/Q$ ratio. Relinquishing the floor and ceiling functions allows us to obtain a crude comparison between the lower and upper limits of Expression \ref{eq:boundaryConditionsForM}. More concretely, we are able to verify that these limits differ by a $\log_{2}{|S_{i}|}$ factor. This means that the system, besides requiring $\lceil \log_{2}{|S_{i}|} \rceil$ bits for register $\ket{x}$, can still employ an additional $\log_{2}{|S_{i}|}$ bits to encode $g$'s output and in the process still perform better than its classical counterpart.

\subsubsection{On the growth of $g$'s output \label{sec:onTheGrowthOfG'sOutput}}

Consider a production system with a constant branching factor where a set of productions is applied then there will exist a total of $|R|^{d}$ possible tree paths at depth level $d$, who will require $\lceil \log_{2}{|R|^{d}} \rceil$ bits for an adequate encoding. Clearly, if $\lceil \log_{2}{|R|^{d}} \rceil  \leq \lceil \log_{2}{|S_{i}|} \rceil$ then $g$'s output can encode the sequence of productions applied. If this is not the case we may opt to encode an unspecified number of productions applied according to some previously chosen strategy. As a consequence, our proposal may be more appropriate when dealing with large $S_{i}$ sets since this would automatically imply that we would have at our disposition a large set of working bits. Even if this is not the case it is still possible to employ as initial states the set of nodes that can be found at a depth $d$ which, as previously mentioned,  typically grow in an exponential fashion.

\subsubsection{Comparison with an hierarchical search model \label{sec:comparisonWithAnHierarchicalSearchModel}}

Finally, it is important to compare our quantum production system performance against a similar proposal described in \cite{tarrataca2010}. In their work the authors also employ Grover's algorithm alongside an hierarchical search oracle where a superposition consisting of all possible paths up to depth-level $d$ is evaluated. The authors chose to build a binary string encoding in a logical fashion the sequence of actions taken. As a consequence a total of $p_{1} = d \times \lceil log_{2}{|R|}\rceil$ bits is required. This approach contrasts with the $p_{2} = \lceil \log_{2}{|R|^{d}} \rceil$ bits employed to encode all possible paths. However, if $|R|$ is a power of $2$, then the use of the ceiling functions is no longer required and it is possible to conclude that $p_{1} = p_{2} = p$. For that reason, the number of bits $m$ required by our current proposition will always be bigger than $p$, since $m = n + p + 1$. This implies that the number of Grover iterations to apply in \cite{tarrataca2010}, respectively, $O(\sqrt{2^{p}})$ will also be less than the $O(\sqrt{2^{n+p+1}})$ time required with this method. If a ratio is performed between both procedures then we are able to verify that they differ by a factor of $\frac{\sqrt{2^{p}}}{\sqrt{2^{n+p+1}}} = \sqrt{\frac{1}{2^{n+1}}}$ in favor of \cite{tarrataca2010}. However, despite loosing in performance terms, our model differs significantly in nature. The authors original proposal focused on exploiting hierarchical search through polynomial time verification of paths, whilst we propose building a generic mechanism for hierarchical search through quantum operators.

\section{Conclusions \label{sec:conclusions}}

In this work we presented an artificial intelligence inspired quantum computational model based on production system theory. Quantum computation is an inherently reversible process and as a consequence the proposed model would also allow for a reversible decision process.  Since production systems share some key characteristics with classical tree search the proposed model also allows for an hierarchical quantum search mechanism. By formalizing the theoretical foundations of our approach we were able to enumerate the reversible and quantum requirements of our model. These requirements enabled us to present a method focusing on the construction of the unitary operator associated with our quantum production system.  We then extended our proposition in order to combine with Grover's algorithm. Doing so allowed us to adequately study the performance of our system and enumerate those cases in which our model outperforms its classical counterpart. Although our proposition is able to compute faster the $\frac{p}{\sqrt{2}} \sqrt{|S_{i}|}$ performance penalty associated with each additional bit required is expensive, favoring the choice of models that rely exclusively on exploiting the class of problems NP through polynomial time verifications and superpositions of all possible paths.

\section*{Acknowledgements}

This work was supported by FCT (INESC-ID multiannual funding) through the PIDDAC Program funds and FCT grant DFRH - SFRH/BD/61846/2009. 

%%%%%%%%%%%%%%%%%%%%%%%%%%%%%%%%%%%%%%%%%%%%%%%%%%%%%%%%%%%%
%       									BIBLIOGRAPHY					         		     %
%%%%%%%%%%%%%%%%%%%%%%%%%%%%%%%%%%%%%%%%%%%%%%%%%%%%%%%%%%%%
\bibliographystyle{spmpsci}      % mathematics and physical sciences
%\bibliography{../../../../../Bibliography/bibliography}

\begin{thebibliography}{10}
\providecommand{\url}[1]{{#1}}
\providecommand{\urlprefix}{URL }
\expandafter\ifx\csname urlstyle\endcsname\relax
  \providecommand{\doi}[1]{DOI~\discretionary{}{}{}#1}\else
  \providecommand{\doi}{DOI~\discretionary{}{}{}\begingroup
  \urlstyle{rm}\Url}\fi

\bibitem{abramsky1999}
Abramsky, S., S., A., Shore, R., Troelstra, A.: Handbook of computability
  theory.
\newblock Elsevier (1999)

\bibitem{anderson1983}
Anderson, J.R.: The Architecture of Cognition.
\newblock Harvard University Press, Cambridge, Massachusetts, USA (1983)

\bibitem{anderson1995}
Anderson, J.R.: Cognitive Psychology and its Implications, fourth edn.
\newblock W. H. Freeman and Company (1995)

\bibitem{bennett1973}
Bennett, C.: Logical reversibility of computation.
\newblock IBM Journal of Research and Development \textbf{17}, 525--532 (1973)

\bibitem{bennett1997}
Bennett, C.H., Bernstein, E., Brassard, G., Vazirani, U.: Strengths and
  weaknesses of quantum computing (1997).
\newblock
  \urlprefix\url{http://www.citebase.org/abstract?id=oai:arXiv.org:quant-ph/97%
01001}

\bibitem{bernstein1993}
Bernstein, E., Vazirani, U.: Quantum complexity theory.
\newblock In: STOC '93: Proceedings of the twenty-fifth annual ACM symposium on
  Theory of computing, pp. 11--20. ACM, New York, NY, USA (1993).
\newblock \doi{http://doi.acm.org/10.1145/167088.167097}

\bibitem{bourbaki2004}
Bourbaki, N.: {Elements of mathematics: theory of sets}.
\newblock No. vol. 1 in Elements of mathematics. Springer (2004).
\newblock \urlprefix\url{http://books.google.pt/books?id=IL-SI67hjI4C}

\bibitem{choi2006}
{Choi}, B., {Korepin}, V.: {Quantum Partial Search of a Database with Several
  Target Items}.
\newblock ArXiv Quantum Physics e-prints  (2006)

\bibitem{chuang1998}
Chuang, I.L., Gershenfeld, N., Kubinec, M.: Experimental implementation of fast
  quantum searching.
\newblock Phys. Rev. Lett. \textbf{80}(15), 3408--3411 (1998).
\newblock \doi{10.1103/PhysRevLett.80.3408}

\bibitem{church1941}
Church, A.: The Calculi of Lambda-Conversion.
\newblock Annals of Mathematics Studies. Princeton University Press (1941)

\bibitem{martin2000}
Davis, M.: The Universal Computer: The Road from Leibniz to Turing.
\newblock Norton (2000)

\bibitem{martin2001}
Davis, M.: Engines of logic: mathematicians and the origin of the computer.
\newblock Norton (2001)

\bibitem{deutsch1985}
Deutsch, D.: Quantum theory, the church-turing principle and the universal
  quantum computer.
\newblock In: Proceedings of the Royal Society of London- Series A,
  Mathematical and Physical Sciences, vol. 400, pp. 97--117 (1985)

\bibitem{deutsch1992}
{Deutsch}, D., {Jozsa}, R.: {Rapid Solution of Problems by Quantum
  Computation}.
\newblock Royal Society of London Proceedings Series A \textbf{439}, 553--558
  (1992)

\bibitem{dirac1939}
Dirac, P.A.M.: A new notation for quantum mechanics.
\newblock In: Proceedings of the Cambridge Philosophical Society, vol.~35, pp.
  416--418 (1939)

\bibitem{dirac1981}
Dirac, P.A.M.: The Principles of Quantum Mechanics - Volume 27 of International
  series of monographs on physics (Oxford, England) Oxford science
  publications.
\newblock Oxford University Press (1981)

\bibitem{ernst1969}
Ernst, G., Newell, A.: GPS: a case study in generality and problem solving.
\newblock Academic Press (1969)

\bibitem{grover1996}
Grover, L.K.: A fast quantum mechanical algorithm for database search.
\newblock In: STOC '96: Proceedings of the twenty-eighth annual ACM symposium
  on Theory of computing, pp. 212--219. ACM, New York, NY, USA (1996).
\newblock \doi{http://doi.acm.org/10.1145/237814.237866}

\bibitem{grover2004a}
Grover, L.K., Radhakrishnan, J.: Is partial quantum search of a database any
  easier? (2004).
\newblock
  \urlprefix\url{http://www.citebase.org/abstract?id=oai:arXiv.org:quant-ph/04%
07122}

\bibitem{gusfield1997}
Gusfield, D.: {Algorithms on strings, trees, and sequences: computer science
  and computational biology}.
\newblock Cambridge University Press (1997).
\newblock \urlprefix\url{http://books.google.pt/books?id=STGlsyqtjYMC}

\bibitem{korepin2006}
Korepin, V., Grover, L.: Simple algorithm for partial quantum search.
\newblock Quantum Information Processing \textbf{5}, 5--10 (2006).
\newblock \urlprefix\url{http://dx.doi.org/10.1007/s11128-005-0004-z}.
\newblock 10.1007/s11128-005-0004-z

\bibitem{korepin2007}
{Korepin}, V.E., {Xu}, Y.: {Hierarchical Quantum Search}.
\newblock International Journal of Modern Physics B \textbf{21}, 5187--5205
  (2007).
\newblock \doi{10.1142/S0217979207038344}

\bibitem{laird1987}
Laird, J.E., Newell, A., Rosenbloom, P.S.: Soar: An architecture for general
  intelligence.
\newblock Artificial Intelligence \textbf{33}(1), 1--64 (1987)

\bibitem{laird1986}
Laird, J.E., Rosenbloom, P.S., Newell, A.: Chunking in soar: The anatomy of a
  general learning mechanism.
\newblock Machine Learning \textbf{1}(1), 11--46 (1986)

\bibitem{luger1993}
Luger, G.F., Stubblefield, W.A.: Artificial Intelligence: Structures and
  Strategies for Complex Problem Solving: Second Edition.
\newblock The Benjamin/Cummings Publishing Company, Inc (1993)

\bibitem{mano2002}
Mano, M., Kime, C.R.: Logic and Computer Design Fundamentals: 2nd Edition.
\newblock Prentice Hall (2002)

\bibitem{markov1954}
Markov, A.: The theory of algorithms.
\newblock National Academy of Sciences, USSR (1954)

\bibitem{newell1963}
Newell, A.: A guide to the general problem-solver program gps-2-2.
\newblock Tech. Rep. RM-3337-PR, RAND Corporation, Santa Monica, CA, USA (1963)

\bibitem{newell1959}
Newell, A., Shaw, J., Simon, H.A.: Report on a general problem-solving program.
\newblock In: Proceedings of the International Conference on Information
  Processing, pp. 256--264 (1959)

\bibitem{newell1972}
Newell, A., Simon, H.A.: Human problem solving, 1 edn.
\newblock Prentice Hall (1972)

\bibitem{nielsen2000}
Nielsen, M.A., Chuang, I.L.: Quantum Computation and Quantum Information.
\newblock Cambridge University Press (2000)

\bibitem{post1943}
Post, E.: Formal reductions of the general combinatorial problem.
\newblock American Journal of Mathematics \textbf{65}, 197--268 (1943)

\bibitem{shor1994}
Shor, P.: Algorithms for quantum computation: discrete logarithms and
  factoring.
\newblock In: Proceedings 35th Annual Symposium on Foundations of Computer
  Science, pp. 124--134 (1994).
\newblock \doi{10.1109/SFCS.1994.365700}

\bibitem{tarrataca2010}
Tarrataca, L., Wichert, A.: Tree search and quantum computation.
\newblock Quantum Information Processing pp. 1--26 (2010).
\newblock \urlprefix\url{http://dx.doi.org/10.1007/s11128-010-0212-z}.
\newblock 10.1007/s11128-010-0212-z

\bibitem{tarrataca2011}
Tarrataca, L., Wichert, A.: Can quantum entanglement detection schemes improve
  search?
\newblock Quantum Information Processing pp. 1--8 (2011).
\newblock \urlprefix\url{http://dx.doi.org/10.1007/s11128-011-0231-4}.
\newblock 10.1007/s11128-011-0231-4

\bibitem{toffoli1980a}
Toffoli, T.: Reversible computing.
\newblock In: Proceedings of the 7th Colloquium on Automata, Languages and
  Programming, pp. 632--644. Springer-Verlag, London, UK (1980)

\bibitem{toffoli1980b}
Toffoli, T.: Reversible computing.
\newblock Tech. rep., Massschusetts Institute of Technology, Laboratory for
  Computer Science (1980)

\bibitem{turing1936}
Turing, A.: On computable numbers, with an application to the
  entscheidungsproblem.
\newblock In: Proceedings of the London Mathematical Society, vol.~2, pp.
  260--265 (1936)

\bibitem{turing1950}
Turing, A.: Computing machinery and intelligence.
\newblock Mind \textbf{59}, 433--460 (1950)

\bibitem{winston1992}
Winston, P.H.: Artificial Intelligence (Third Edition).
\newblock Addison-Wesley (1992)

\bibitem{ying2010}
Ying, M.: Quantum computation, quantum theory and ai.
\newblock Artificial Intelligence \textbf{174}, 162--176 (2010)

\end{thebibliography}

\end{document}